\documentclass{llncs}
\usepackage{comment}

\usepackage{mathtools}
\usepackage{multirow}
\usepackage{graphicx}
\usepackage{subfigure}
\usepackage{wrapfig}
\usepackage{adjustbox}
\usepackage{makecell}
\usepackage{url}
\RequirePackage[bookmarks,
                urlcolor=black,
                citecolor=black,
                linkcolor=black,
                pagecolor=black,
                colorlinks,
                hyperfigures]{hyperref}
\usepackage{csquotes}
\usepackage{todonotes}
\usepackage[noend,linesnumbered]{algorithm2e}
\usepackage{placeins}
\usepackage[inline]{enumitem}
\usepackage{array}
\usepackage{listings}
\usepackage{pgfplots}
\usepackage{pgfplotstable}
\usepackage[gen]{eurosym}
\usepackage{amsfonts}
\usepackage{amsmath}

\DeclareRobustCommand{\circled}[1]{\tikz[baseline=(char.base)]{\node[shape=circle,draw,font=\small\sffamily,inner sep=0pt,minimum size=4.5mm] (char) {#1};}}

\begin{document}

\mainmatter  
\title{Sustainability Through Cognition Aware Safety Systems - Next Level Human-Machine-Interaction}

\author{Juergen Mangler$^1$, Konrad Diwold$^{6,7}$, Dieter Etz$^3$, Stefanie Rinderle-Ma$^1$, Alois Ferscha$^{2,6}$, Gerald Reiner$^4$, Wolfgang Kastner$^3$, Sebastien Bougain$^5$, Christoph Pollak$^5$, Michael Haslgrübler$^6$}
\institute{$^1$~Technical University of Munich, Faculty of Computer Science, Munich, Germany\\
\email{\{firstname.lastname\}@tum.de}\\
$^2$~Johannes Kepler University Linz, Faculty of Computer Science, Linz, Austria\\
\email{\{lastname\}@pervasive.jku.at}\\
$^3$~Technical University of Vienna, Faculty of Computer Science, Vienna, Austria\\
\email{\{firstname.lastname\}@tuwien.ac.at}\\
$^4$~Vienna University of Economics and Business, Department of Information Systems and Operations Management, Institute for Production Management\\
\email{\{Gerald.Reiner\}@wu.ac.at}\\
$^5$~Center for Digital Production GmbH, Vienna, Austria\\
\email{\{firstname.lastname\}@acdp.at}\\
$^6$~Pro2Future GmbH, Linz, Austria\\
\email{\{firstname.lastname\}@pro2future.at}\\
$^7$~Graz University of Technology, Department of Electrical Engineering, Institute of technical Informatics, Graz, Austria\\
}
\maketitle

\begin{abstract}
Industrial Safety deals with the physical integrity of humans, machines and the environment when they interact during production scenarios. Industrial
Safety is subject to a rigorous certification process that leads to inflexible settings, in which all changes are forbidden. With the progressing introduction
of smart robotics and smart machinery to the factory floor, combined with an increasing shortage of skilled workers, it becomes imperative that
safety scenarios incorporate a flexible handling of the boundary between humans, machines and the environment. In order to increase the well-being of workers, reduce
accidents, and compensate for different skill sets, the configuration of machines and the factory floor should be dynamically adapted, while still enforcing
functional safety requirements. The contribution of this paper is as follows: (1) We present a set of three scenarios, and discuss how  industrial safety mechanisms could be augmented through dynamic changes to the work environment in order to decrease potential accidents, and thus increase productivity. (2) We introduce the concept of a Cognition Aware Safety System (CASS) and its architecture. The idea behind CASS is to integrate AI based reasoning about human load, stress, and attention with AI based selection of actions to avoid the triggering of safety stops.
(3) And finally, we will describe the required performance measurement dimensions for a quantitative performance measurement model to enable a comprehensive (triple bottom line) impact assessment of CASS. Additionally we introduce a detailed guideline for expert interviews to explore the feasibility of the approach for given scenarios.

\end{abstract}

\section{Introduction}
\label{sec:intro}

Today digitalization increasingly permeates all aspects of production and has to balance the interactions between machines, humans, software and the environment. This development is referred to as the fourth industrial revolution or Industry 4.0 (I4.0) in short~\cite{lasi2014industry}, and leads to smart manufacturing systems with high degrees of autonomy and automation. 

While full automation, with no involvement of humans seems feasible for some production use-cases, the participation of humans in a production process is and will be ubiquitous in production environments of the future. What changes though is the quality of the integration of humans in the production process~\cite{vuori2019digitalization}, as the forms of interaction between human operators and production systems are becoming more complex, meaningful and autonomous, which results in a transition of humans from 'mere operators' of machines towards strategic decision makers in flexible production environment~\cite{gorecky2014human}.

For humans this development has a two-fold outcome: On the one hand studies suggested that digitalization enables greater productivity and efficiency within production~\cite{vuori2019digitalization}, as it allows workers to focus on complex tasks that require human knowledge, as simple (often repetitive tasks) can be achieved automatically~\cite{cijan2019digitalization}. On the other hand the information flow within manufacturing processes is continually increasing in complexity and speed. A human worker operating processes usually deals with many different types of subsystems and tasks and interacts with different types and levels of automation. This increases the cognitive burden, often referred to as stress \cite{yamamoto2019stress}, on an operator and can consequently lead to mistakes (due to inattentiveness or inexperience) resulting in a decrease of production quality \cite{kolus2018} or in a worst case scenario accidents, which are mostly caused by humans as far back as 1920s \cite{goetschoccupational}. 

\begin{figure}
  \centering
  \includegraphics[width=1\textwidth]{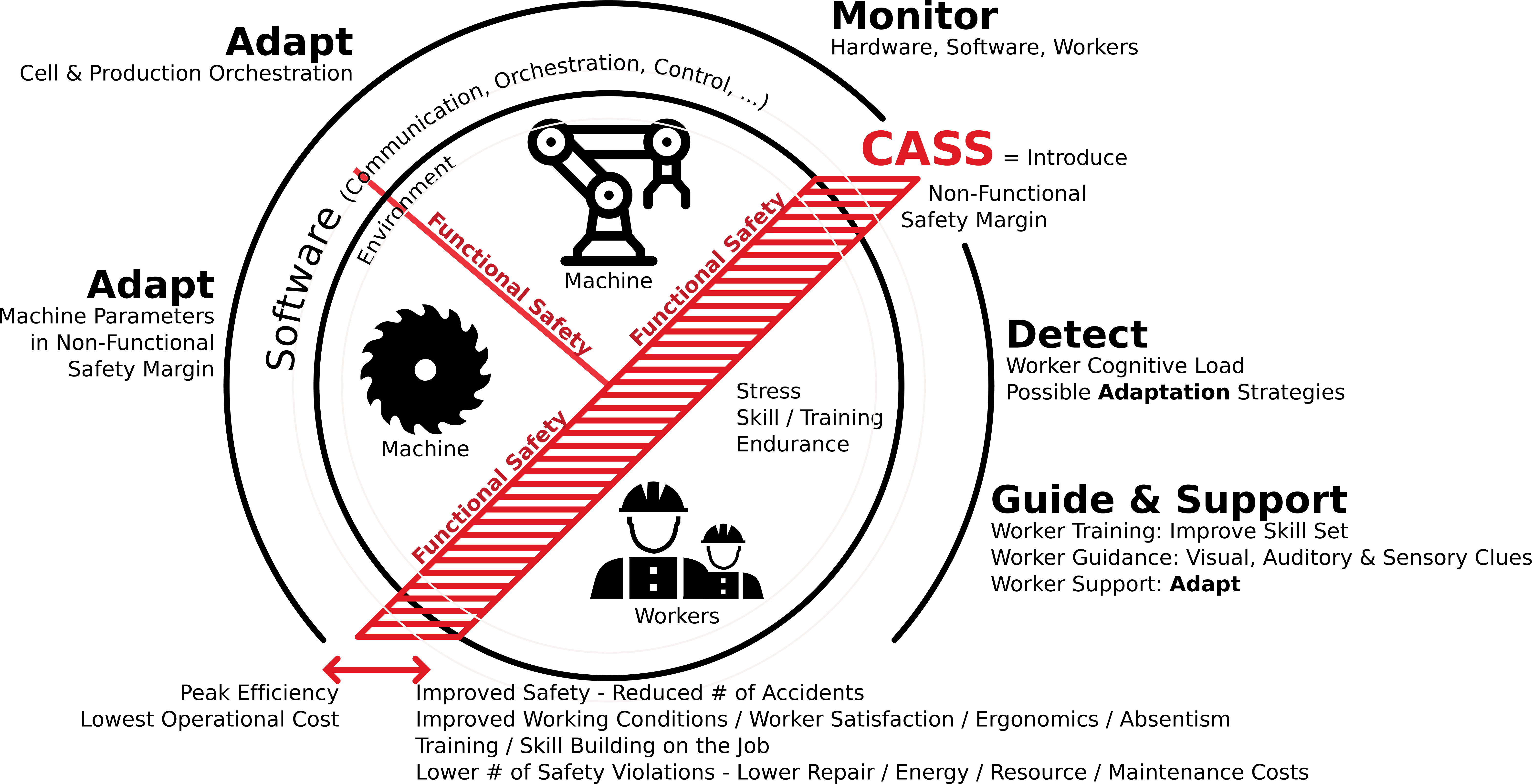}
  \caption{Cognitive Aware Safety System Vision}
  \label{fig:strategy}
\end{figure}

In a modern I4.0 \cite{lasi2014industry} setting, the following aspects have to be balanced, cf. Fig. \ref{fig:strategy}, and optimized: (i) Functional Safety (all measures to avoid all harm to humans) as defined in regulation  \cite{european_union_directive_2006} and norms \cite{iec_61508_2011}, (ii) Cell \& Production Orchestration \cite{pauker2018centurio} while also considering varying levels of human performance which is modulated by (i) frequent changes in terms of concentration and fatigue  and (ii) more slowly changing factors like skill level or training level.

So far the balancing problem between functional safety and production orchestration has not been sufficiently addressed as currently state of the art systems just trigger safety violations and therefore help to avoiding harm to humans and the environment. However the cost of these safety violations is not considered: (1) time lost and energy wasted while restarting the production, (2) resource costs due to erroneously affected parts (energy, raw material), (3) repair and maintenance cost, and (4) potential penalties due to missed deadlines.

The contribution of this position paper is therefore five-fold: (1) We elaborate why and how cognitive load might affect worker behaviour through an extensive literature review. (2) We then present a set of three real-world scenarios from industrial partners, including a preliminary analysis how they are impacted by worker behaviour. (3) We discuss for each scenario, how a diverse group of workers (different skill sets, different cognitive loads, ...) can be supported to reduced the number of safety violations, and thus potentially reduce cost and improve worker satisfaction, without touching existing laws and regulations regarding functional safety. (4) We then introduce the concept of Cognition Aware Safety System (CASS) (see Fig.~\ref{fig:strategy}). CASS is envisioned to be a software architecture utilizing (a) artificial intelligence techniques to adapt system parameters, (b) visual, auditory and sensory worker guidance techniques, (c) worker training through enhancing real-world scenarios with Augmented/Virtual Reality (AR/VR). 
(5) And finally we lay out how we plan to assess the impact of CASS on a given production scenario, including required quantitative performance measurement dimensions, as well as a guideline to conduct qualitative exploratory empirical research based on expert interviews.

The paper is structured as follows: Sec. \ref{Sec:cognition} describes the challenges of dealing with cognition loads in manufacturing. Section \ref{Sec:safety} introduces how industrial safety systems work, certification challenges, and how CASS might help improve the situation. We then introduce three detailed real-world scenarios from company partners in Sec. \ref{Sec:scenarios}. This section is not only used as the basis for describing the potential impact of CASS, but also contains detailed preliminary information about potential frequency of safety errors as well as potential savings that can occur through CASS.
In Sec. \ref{Sec:adapt} we describe the CASS architecture in detail: (1) how to we plan to track human behavior, (2) how to we plan to detect anomalies, and (3) how do we plan to act.

In order to evaluate our approach, in Sec. \ref{sec:eval} we lay out a detailed strategy for quantitative as well as qualitative evaluation. Additionally, the appendix to this paper provides a detailed guideline for expert interviews that will be conducted with our company partners to asses the feasibility and usefulness of the approach. Finally, Sec. \ref{sec:conclusion} discusses the next steps that will be taken in order to realize CASS.

\section{Cognition in manufacturing} \label{Sec:cognition}

The increased integration of technology such as robotics, artificial intelligence, or the internet of things into shop floors and their transition towards cyber-physical systems, is leading to a general change of required human labor and work organization in general~\cite{bonekamp2015consequences}, as work shifts from low-skilled tasks towards high-skilled complex jobs \cite{yamamoto2019stress}. On the workers’ side these jobs  require continuous learning, with a worker becoming “the most flexible component'' in a production system, who need to strategically intervene in the autonomously organized production system~\cite{gorecky2014human}. The work environment itself becomes “a highly interconnected network of people, technology and business units''~\cite{adriaensen2019can}. Advanced manufacturing processes and environments themselves can generate new occupational health safety (OHS) risks, which require novel forms of analysis~\cite{fernandez2015analysis}. On the one hand these risks concern novel material used in the context of advanced production such as nanoparticles~\cite{fabiano2019safety}. According to the European agency for safety and health, the increase in and design of complex human-machine interaction bears another risk as an increase in complexity of automation technologies in combination with poorly designed human-machine interfaces can yield to increased mental stress on workers \cite{hauke2018proactive,yamamoto2019stress}. Additionally, system failure might lead to more severe consequences than failures in less automated systems~\cite{onnasch2014human}. This is corroborated by recent findings of Neumann et al.~\cite{neumann2021industry}, which states that humans have been broadly ignored in the design of novel automation concepts and call for the systematic integration of human factors into future manufacturing research and development. In another analysis Kolos et al.~\cite{kolus2018} shows the quality related risk factor, such as difficulty and load, of humans in production and argues for addressing these.

An important factor in operating an automation system is the current cognitive load of a worker during task operation~\cite{carvalho2020cognitive,mattsson2020forming,thorvald2019development}. Cognitive load is defined as “the cognitive effort made by a person to understand and perform his/her task”~\cite{sweller1988cognitive}. According to Sweller, cognitive load can be distinguished into three types: Intrinsic cognitive load constitutes the cognitive load which is inherent to a certain task, and is defined by a task’s intrinsic complexity, but varies based on the experience level of the person performing the task. Extraneous cognitive load is caused by the need of processing information, collaboration and the mental effort required to coordinate a certain task effort. As it is ineffective it should be reduced as much as possible. Germane cognitive load constitutes learning efforts thus storing information in the long-term memory. In the context of learning (e.g., to operate an automation system), germane cognitive load should be stimulated by the environment in order to enable a learning process of the perceived information and it’s mere processing. Cognitive load theory aims for the development of new learning methods which facilitate the efficient use of cognitive abilities to optimize knowledge and skill gain in humans~\cite{paas2003cognitive}. The impact of cognitive load on task performance has been studied in various domains such as driving (e.g., ~\cite{engstrom2017effects,kun2011cognitive}) or medical applications~\cite{dias2018systematic}. In the context of manufacturing intrinsic and extraneous loads are thought to be of highest relevance~\cite{carvalho2020cognitive}. The shifts from low-skilled tasks towards high-skilled complex jobs resulting within environments with high automation levels results in an increase in intrinsic task difficulty (thus an increase in intrinsic cognitive load) and high coordination efforts for successful task execution (thus an increase in extraneous cognitive load)~\cite{carvalho2020cognitive}. In the context of assembly quality defects result from product errors (wrongly assembled components) or process errors (assembly process incomplete/wrong), which both can be the result of excessive cognitive load on the human operator’s side~\cite{thorvald2019development}. As shown in~\cite{kumar2016measurement}  the modalities used for human machine interaction can have a significant impact on cognitive load, with visual instructions leading to less brain activity and hence less cognitive load caused by the machine than an auditory counterpart. 
A number of techniques aim for the assessment of cognitive load. On the one hand these methods concern subjective rating methods, which aim for assessing cognitive load via questionnaires and are based on the assumption that people are able to assess and report expended cognitive effort~\cite{thorvald2019development}. Most prominent subjective rating methods are the NASA-Task Load Index~\cite{paas2003cognitive}, Subjective Workload Assessment Technique~\cite{rubio2004evaluation} and the Subjective Workload Dominance Technique \cite{Vidullch1991,stanton2005driver}. Physiological methods on the other hand assume that levels of cognitive load and changes therein are reflected in physiological variables \cite{sweller1998cognitive}. A number of physiological indicators have been identified and used in the context of establishing a person's cognitive load. These variables include heart activity~\cite{akselrod1981power}, brain activity~\cite{kumar2016measurement}, eye activity~\cite{ahlstrom2006using}, skin conductance/response~\cite{mehler2009impact} and  cortisol  levels~\cite{carrasco2003van}. While establishing physiological “cognitive load” indicators were initially restricted to laboratory settings, the developments in the context of wearable and IoT has lead to inexpensive and less invasive means of establishing cognitive load in real application environments. Das et al.~\cite{das2014cognitive} developed an inexpensive EEG device to assess a person’s cognitive load by measuring brain activity. A study by Gjoreski et al.~\cite{gjoreski2018my} demonstrated how wearable sensing devices in combination with machine learning models can be used to assess a person’s cognitive load and compared the results with subjective rating methods.

\section{Functional Safety} \label{Sec:safety}

One part of modern machinery, which is compulsory by law, is a functional
safety system. It is an active system with the purpose to detect potentially
dangerous conditions resulting in the activation of a protective or corrective
device or mechanism to prevent hazardous events arising \cite{iec_safety_2005}
\cite{european_union_directive_2006}.

IEC~61508, one of the most important functional safety standards, covers safety
management, system and hardware design, software design, production, and
operation of safety critical systems. It defines safety as the ``\textelp{}
freedom from unacceptable risk of physical injury or of damage to the health of
people, either directly, or indirectly as a result of damage to property or to
the environment.'' Therefore, the main elements to consider in a safety related
application are: humans, hardware (e.g. machinery,
electrical/electronic/programmable electronic systems), software, and
environment \cite{iec_61508_2011}.  Current law requires a machine designer to
assess the risks of a machine and define counter measures against each
potential hazard. For each identified hazard, a safety function has to be
speciﬁed, documented, and verified, including all Safety-Related Parts of
Control Systems (SRP/CS) such as sensors, application logic, actuators, and the
connections in between \cite{iso_safety_2010}.

Smart manufacturing facilities, in the context of Industry 4.0, introduce new
requirements on machinery, such as flexibility, interoperability, and dynamic
reconfiguration, just to name a few.  These new requirements and their
implementation, result in a high increase of complexity. To overcome these
difficulties, systems are required which assist the operators to perform their
daily tasks and enhance the efficiency of the production line
\cite{etz_flexible_2020}.  One example of this is the safety protection area
around a robot cell. The simplest but also inflexible solution to protect
humans from potential hazards trough a moving robot was predominantly done by
the use of physical barriers, such as fencing and gates. These barriers fulfill
the safety requirements, but introduce new issues regarding flexibility and
efficiency especially in a setup where the operator has to enter the safety
protection area \cite{Augustsson2014}.  In recent years, several safety
certified optical and radar solutions have been launched to the market that are
able to replace physical barriers. These solutions, such as laser scanners,
light curtains, safe camera systems, and safe radar systems, are capable of
reliably monitor a safety protection area \cite{Zlatanski2018}. In this context,
CASS aims to help that an operator's cognitive state (i.e. inattention) does
not reduce the efficiency of the machine by preventing a production stop due to
an unintended activation of functional safety measures~\cite{Halme2018}.
Furthermore, these devices are configurable and allow free movement in the
safety protection area when allowed. This flexible solution comes with one
drawback: the borders of the safety protection area are invisible. Therefore,
triggering the safety measures by unintended entering of the safety protection
area because of inattentiveness, inattention, or distraction could easily
happen.  In this context, CASS aims to help that an operator's cognitive state
(i.e. inattention) does not reduce the efficiency of the machine by preventing
a production stop due to an unintended activation of functional safety
measures.  This improvement will be achieved by the introduction of a safety
margin prior to triggering of the functional safety system by monitoring and
adapting to cognitive state. Thus CASS can give guidance, assistance, and
advice to the operator, as well as adapt the machine configuration, in order
that no cognitive overload occurs and therefore mistakes are being prevented.

\section{Application Scenarios} \label{Sec:scenarios}

In order to illustrate how human load, stress, attention, but also skill and experience influence production scenarios, the following subsections will provide three real-world scenarios from industrial partners. The goal is is exemplify how these existing production situations could be made more efficient, more safe and more sustainable. In particular we provide examples how Cognition Aware Safety may address quality issues -  \ref{ssec:app1}, safety issues - \ref{ssec:app2} and flexibility - \ref{ssec:app3}.

\subsection{Tool and Work-Piece Loading Support for Flexible Manufacturing Systems}
\label{ssec:app1}

Company "A" is a mid sized contract manufacturer of high precision metal parts for industrial applications. The typical lot size of the orders ranges from 5 to 1000. The company has approximately 150 employees. The size of the produced parts is in the range between several centimeters up to approximately half a meter. The credo of company "A" is to be able to achieve constant manufacturing cost with respect to the lot size, an expectation which is increasingly expressed by customers. Customers expect to order based on immediate demand and are reluctant to commit to long term/high volume procurement contracts. Typical expected delivery times range from 2-3 days to 4 weeks. 

Machining is carried out in milling centers being integrated in a flexible manufacturing system (FMS), consisting of manual loading stations, forklifts, robots, conveyors, as well as buffers. The interface between the shop floor personnel and the FMS is a loading station, where the operators have to perform the following tasks: 

\begin{itemize}
   
   \item The correct tombstone (part between pallet and vice) has to be selected by the operator.
   
   \item The vice has to be mounted on the tombstone.
   
   \item The correct fixture has to be selected and mounted to the vice.
   
   \item The raw part has to correctly positioned.
   
   \item The raw part has to be clamped with the appropriate force.
   
\end{itemize}   
   
After loading, the pallet is transported by conveyor and fork-lifting systems to internal buffer positions and to the milling centers. The pallets are loaded/unload by robots into the milling centers. Depending on the product, it is possible that a pallet might enter multiple milling centers to carry out multi-step machining (different milling centers might have different properties, such as spindle power or precision). After the finalisation of the last operation, that pallet is transported back to the loading station, where the 
parts are removed from the clamping devices. 

Depending on the particular quality assurance procedures (which vary from part to part), a visual inspection or a measurement of the key features is conducted, before a part is made ready for delivery. Again the loading station serves as the interface between the shop floor personnel and the FMS:

\begin{itemize}

  \item The visual inspection at the end has to yield a correct result (detect scratches, burrs, milling errors).
  
  \item Depending on the order (tolerances) and the part (features), manual measurement of key features has to be conducted and documented.
  
  \item Depending on the order, a subset of the parts has to be selected for machine measurement  (e.g. with a coordinate measurement system).

\end{itemize}    

Besides the loading and unloading of the parts, operators have to provide the FMS with the correct sets of cutting tools for the machining tasks, for this purpose, the following tasks have to be carried out by the shop floor personnel: 

\begin{itemize}

    \item Tool holders have QR codes, tools are associated with the containing tool holder.
    
    \item Tools have to be prepared, including shape detection and high precision measurement (1µm) of the tool dimensions
    
    \item Tools are stored in a central tool repository.
    
    \item Tools are are automatically transported on a separate conveyor system and inserted into the machines by robot.
    
\end{itemize}

This high degree of automation ensures that the decoupling of manufacturing cost and order size is maximized, as (1) the human involvement is minimized, (2) tools and raw material preparation is externalized, and thus the setup time is minimized, and (3) machine availability is maximized and independent of the order sequence. However still errors exist and need to be addressed, which problems and how CASS can avoids them is presented in \ref{tab:casserrorsa}. Frequency describes the share of this problem on NOK parts. Impact denotes the time lost per part in relation to the part cycle time.

\iftrue
\begin{table}
  \caption{In order to eliminate the below error sources, workers can be supported by CASS based on cognitive load and skill level by providing the following functions for Company "A": (a) Overlays indicating specific orientation/position/force/key feature information for the case to the operator, (b) Potentially enforce additional setup/quality checks, (c) Balance work (on-the-fly) between loading stations}
  \label{tab:casserrorsa}

  \begin{center}
    \begin{adjustbox}{max width=\textwidth}
    \begin{tabular}{lccc}
    Problem & Frequency & Impact & Solution \\\hline
   Errors during clamping: orientation of raw part &20\% & 100\%& a\\
   Errors during clamping: position of raw part &25\% & 100\%& a \\
   Errors during clamping: use of inappropriate fixture or vice &25\% & 50\%& a \\
   Errors during clamping: inappropriate position of clamping element (constraining contour) &25\% & >100\%& a \\
   Errors during clamping: inappropriate clamping force &3\% & 10\%& a \\
   Visual inspection error: &1\% & 0\% - >100\%& a \\

   Manual measurement errors: wrong key features, wrong documentation &1\% & 0\% - >100\%& - \\
   Coordinate measurement problem: part selection bias &1\% & 0\% - >100\%& a,c \\
   Coordinate measurement problem: inappropriate selection of features &1\% & 0\% - >100\%& - \\
   Coordinate measurement problem: insufficient part cleaning &1\% & 0\% - >100\%& b,c \\
   Coordinate measurement problem: positioning error &1\% & 0\% - >100\%& a \\

    \makecell[l]{
   Errors during tool preparation: wrong tool in tool-holder \\(if the tool has a similar contour, and the system can not detect the deviation)} &1\% & 100\%& b \\
    \end{tabular}
    \end{adjustbox}
  \end{center}
\end{table}
\fi

The goal in this scenario is to (1) track and (2) minimize errors that might occur during the interaction of humans and the FMS. Humans can be supported by CASS in many aspects by (1) introducing additional work steps to ensure error minimization, time loss and quality of parts, and the (2) overlay useful information to reduce stress and improve the well-being of workers.

Potential savings and how they could be achieved are summarized in Tab. \ref{tab:casserrorsa}.

\subsection{High Voltage Test Adapter for Electric Device Inspection}
\label{ssec:app2}

Company "B" is a medium-sized electronics manufacturer of devices for use in control cabinets and switchgears in machine control and systems engineering. The company has approximately 75 employees. 

Production of the electronic control devices is carried out in a production facility consisting of Printed Circuit Board (PCB) manufacturing, automatic and manual PCB component assembly, mechanical assembly, and testing.
Some of the tests during the final inspection are high voltage tests where up to 6000V are applied to the Device Under Test (DUT). Such tests have to be carried out within a test adapter with special safety measures.

The interface between humans operating the testing and the test setup is the test adapter which is mounted on a table. In order to test the proper functionality of the devices, the operator has to carry out the following steps: 

\begin{itemize}
   
   \item The test program has to be selected by human.
   
   \item The DUT has to correctly positioned on the base plate.
   
   \item A fixture with needle adapter has to be positioned on the DUT.
   
   \item The test setup has to be fixated by pulling down a protective hood.
   
\end{itemize}   

The testing of high voltage electric equipment with logical functions (e.g. relays) is usually done on a 100 per cent Level, i.e. every part is tested before being dispatched. As a consequence, the testing represents a significant cost factor. The test benches have to be designed for every product individually (because of different geometries, different numbers of contacts etc.).

From as safety perspective, the housing is required. The housing is on top of the DUT, in order to protect the person that tests the device (operator) is safe, as soon as the testing starts.
After the housing is on top of the DUT, the operator presses a button, and initiates an automatic test cycle that initiates different input variations and confirms that the output of the DUT is correct.

The operator also visually inspects the device and its functionality during the tests. After finishing the tests the serial number is flashed and - after removing the housing - a QR code is put on the device itself.

This situation could be improved when the housing is no longer necessary.
A situation where multiple devices can be tested in a single station, would
require that a user can remove and insert individual DUTs while other DUTs are under test. Touching the wrong DUT from a safety perspective should be impossible, so a safe distance of the user is a functional safety requirement. In case safety is violated, the currently running test has to be repeated, the throughput of the testing station is decreasing.

In this scenario CASS should provide different levels of visual guidance to stop users from touching a DUT which is currently in a high-voltage test. Tailored warnings for high stress situations, inexperienced users, and different voltage levels will lead to improved throughput.

Such a solution would allow the following accelerations and/or cost savings of the texting procedures because a housing is no longer needed:

\begin{itemize}
    \item reduced design and material cost for the testing device (reduced number of customised mechanical parts) 
    \item accelerated deployment of a new testing procedure (facilitated refinement \& troubleshooting due to "expert mode" and "standard/serial mode") 
    \item accelerated part handling because adding and removing the housing is no longer necessary 
    \item parallelization of activities: manipulation on the DUT can be continued after the start of the low-voltage testing sequence. 
\end{itemize}

This has benefits in terms of better working conditions and faster test runs which consequently will lead to shorter throughput times and lower testing costs.

For this use case it is much harder to collect information about potential savings (see Tab. \ref{tab:scenar2}). As the integration of CASS will lead to a full redesign of the testing procedure, the time savings a potentially huge, but can not be precisely quantified.

\begin{table}[ht]
    \centering
    \caption{Frequency is missing because the improvements are applicable for all cases. Impact denotes the time saved per unit.}
    \label{tab:scenar2}
    
    \begin{adjustbox}{max width=\textwidth}
    
      \begin{tabular}{lccc}
      Problem & Impact \\\hline
      Time consuming housing installation & 25\% \\
      Time consuming housing removal & 25\% \\ 
      Throughput (only single parts tested) & - \\
      Housing design effort & - \\
      Adapter design effort & - \\
      \end{tabular}
    \end{adjustbox}
    
\end{table}

\subsection{Automated Work Shop}
\label{ssec:app3}

Company "C" is a manufacturer specialised for the production of small lot sizes of high precision metal parts for industrial applications. The infrastructure of the company consists of 2-3 turning machines, 2-3 milling centers and a coordinate measurement machine (CMM). The company  employs about 20 employees in the following functions:  sales and management (~ 5 persons), work preparation (1-2 persons and machine operators (~13 persons). The size of the manufactured parts is in the range of typically >5 - 300 mm. The usual order size is 1 - 20 pieces and customers expect delivery times of 1 to 3 weeks.

Due to the high variability of the work pieces, 
\begin{itemize}
  
  \item for each ordered part the whole production life-cycle has to be covered: from programming of the machines and defining the manufacturing environment (machine, clamping elements, tools), to material sourcing, setup, and planning of quality assurance steps.
  
  \item the implementation of automation is extremely difficult or is not cost efficient, as for example robot handling typically requires custom grippers and programming. 
  
\end{itemize}

As a result Company "C" usually operate the shop-floor in a one-shift mode, which allows to cope with the high need of internal communication. Moreover a two- or three shift mode would cause over-proportional additional labour cost. 
The downside to this approach is, that the expensive infrastructure is only used 1/3 of the time.

The combination of CPPS, "soft" and flexible safety systems, and cognitive support for the machine operators will allow Company "C" to implement a three-shift mode: 

\begin{itemize}
    \item Shift 1: "handwork", lot size 1-2, full staffing: The company operates "as is", i.e., prototyping (manufacturing of new parts) is done in the conventional way with expert-evaluation of the incoming orders, discussions with customers on manufacturability and cost-saving approaches, definition of manufacturing set-up, (CAM)NC programming, human-surveilled machining, and subsequent quality inspection either manually of in a CMM.
    
    \item Shift 2: "semi-automated", lot sizes 3-20, at 20-30\% staffing: Known parts, that have been already produced before in shift 1 setting, can be produced in shift 2 in a semi-automated manner. Semi automated in this context means: the pallets still have to be manually loaded, but a robot (with existing gripper configuration and programming) can automatically load the machine. Intra-logistics, i.e., moving the parts between machines, again is a manual task.
    This requires that humans enter the shopfloor while robots are performing loading and unloading of the machine tools. Consequently, safety measures have to ensure that no collision between humans and robots are possible and that humans are warned before they enter a safety zone, to avoid emergency stops.
    
    \item Shift 3: "fully automated" ghost-shifts, lot sizes > 20, at minimal staffing: Larger Lot sizes should be manufactured in a fully automated manner, comparable to the mode of execution in a flexible manufacturing system: parts are pre-loaded on a central raw part provisioning station, from where automated guided vehicles (AGVs) carry out the internal transportation to and from the machining centers as well as measurement systems. In this case, a minimum number of employees is on a stand-by-duty. In case of incorrect system behaviour the respective employee has to fix the problem and re-start the manufacturing system. Emergency stops have to be limited to the affected system components which also implies the need for appropriate and easy-to-understand information to operators. 
\end{itemize}

For company "C" CASS is a central element to implement shift 2 \& 3. For shift 2 CASS can minimize down-time and allow for efficient interactions between a diverse set of workers and machines. For shift 3, where a stand-by operator is fixing critical problems, the worker guidance aspect of CASS is crucial, to help him or her to cover the potentially wide array of activities he or she has to perform.

For company "C" the strategic value is two-fold: (1) the infrastructure use is tripled, and (2) a longer planning horizon for future orders (order pipeline) can be achieved.

\subsection{Application Scenario Summary}

These three application scenarios cover a range of issues: quality \ref{ssec:app1}, safety \ref{ssec:app2}, and flexibility \ref{ssec:app3}. For the first scenario CASS can augment the current production, and the potential gains, as depicted in Tab. \ref{tab:casserrorsa}, are very clear. For scenario 2 the gains become more difficult to grasp, as CASS allows for complete (much-requested) redesign of the testing process. This makes Tab. \ref{tab:scenar2} much more generic, but nonetheless the gains might be much bigger than in scenario 2. The same goes for scenario 3, where CASS will allow for a very different way of approaching the production process. As additional shifts are possible, the gains are again potentially huge.

\section{Adaptation Strategy} \label{Sec:adapt}

\begin{figure}[htb]
  \centering
  \includegraphics[width=1\textwidth]{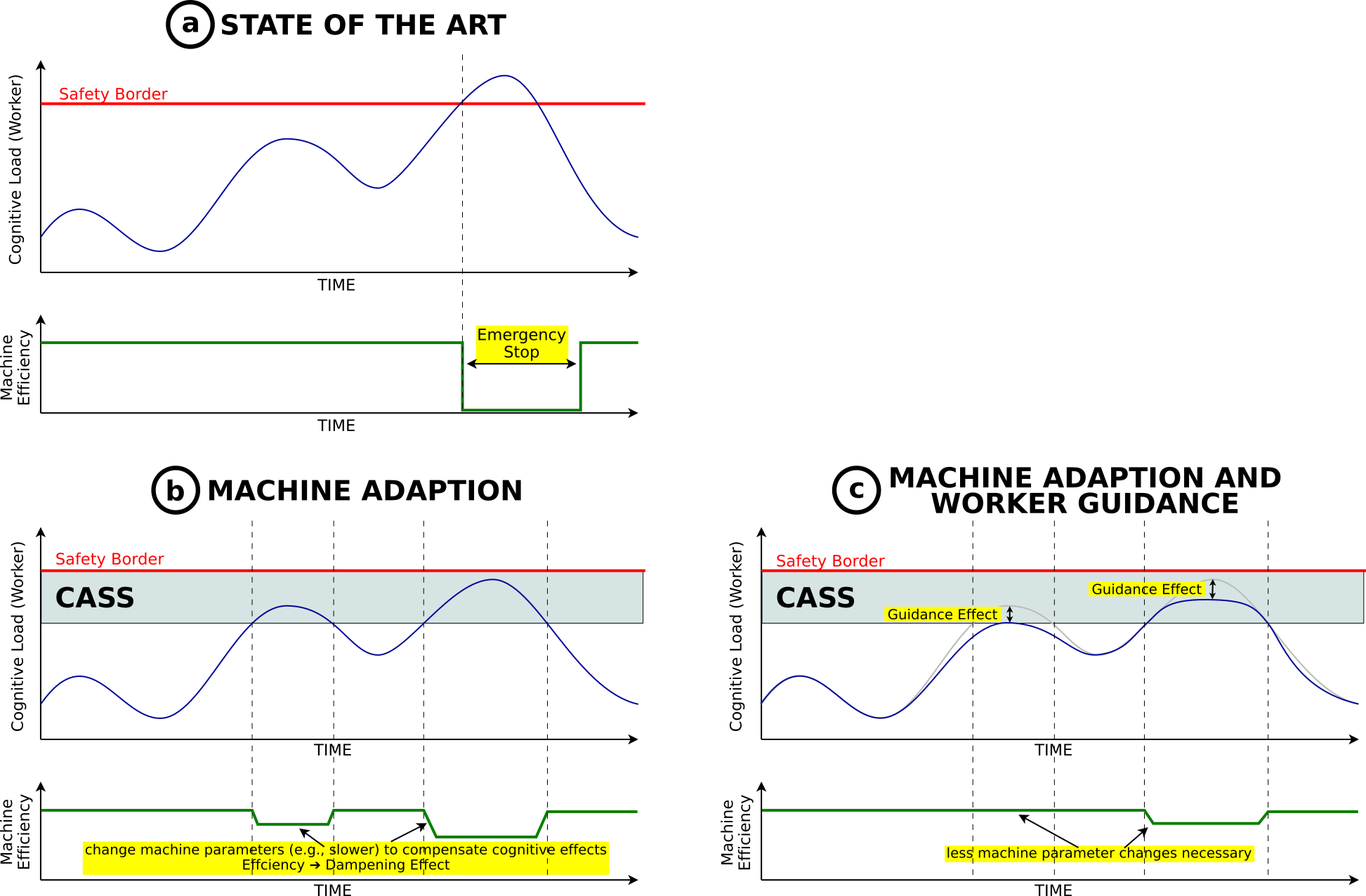}
  \caption{CASS Actions vs. Machining Efficiency}
  \label{fig:adaptation}
\end{figure}

As elaborated in Sec. \ref{Sec:cognition} and Sec. \ref{Sec:safety}, as well as exemplified in Sec. \ref{Sec:scenarios}, CASS will improve the state-of-the-art by reacting to human behavior before a functional safety violation actually occurs. The two pillars to improve machine efficiency are adaptation of machine parameters and worker guidance, as depicted in Fig. \ref{fig:adaptation}. In order to realize CASS we have to define an adaptation strategy that includes (1) how we plan to collect data, (2) the granularity of data that is required for analysis, (3) which potential actions we can take based on analysis results, and (4) how we plan to integrate CASS into a production environment. This culminates into an adaptation strategy, which is the subject of this section.

Defining the adaptation strategy is a design time job. It relies on three things that have to be monitored, partitioned and formalized:

\begin{itemize}
    
    \item Human-machine interactions, e.g., insertion of raw material or tools into a machine.
    
    \item Possible automatic changes to machine parameters that can be made without violating functional safety mechanisms, e.g., slowing down the operation of the a machine, or moving on different paths.
    
    \item Possible automatic changes to the environment that can be made without violating functional safety mechanisms, e.g., expanding safety distances.
    
\end{itemize}

Tracking \textbf{human-machine interactions} is hard. Although many approaches exist to actively track human load, stress, and attention, they are often connected to specialized hardware.  In~\cite{romine2020using} wearables were used to assess a multitude of physiological “cognitive load” indicators and develop a cognitive load tracker. A very recent study of Alam et al.~\cite{alam2020autocognisys} developed  an IoT-based Automatic Cognitive Health Assessment, by combining physiological and physical sensors and using machine learning models to assess cognitive health based on input data. 

\begin{figure}
  \centering
  \includegraphics[width=0.5\textwidth]{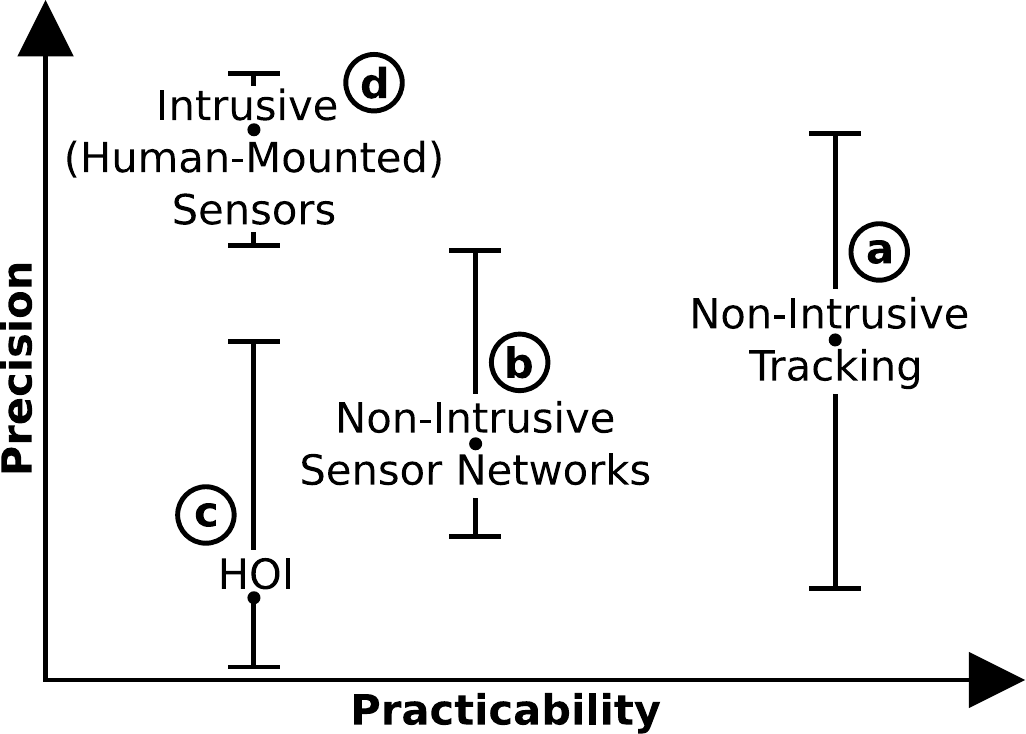}
  \caption{Tracking Human Behavior - Practicability vs. Precision}
  \label{fig:trackingcat}
\end{figure}

While this tracking has the potential precisely monitor human behavior and cognitive load, it seems rather impractical in real-world scenarios, where carrying hardware - including sensors, augmented reality devices, eye-tracking equipment, \ldots{} - (a) are connected to privacy issues, (b) impede free movement of the works, and (c) impose additional management and maintenance issues, e.g. for charging and stewarding devices between different shifts and workers. The perceived balancing between practicability and precision for different approaches is depicted in Fig. \ref{fig:trackingcat}

As for CASS we target wide-spread application in real-world applications even for SMEs, we want to prioritize the following ways of tracking human behavior.

\begin{itemize}
  
  \item[\circled{a}] Implicit detection of human by monitoring their effects and interaction  on and with machines and the environment \cite{pepa2021,Carneiro2012,zhang2018touch}. No additional hardware is required, but we instead purely rely on fine-grained data-collection and data-analysis from machines and already existing sensors in the production environment (e.g. for tracking quality, functional safety, \ldots). All the data collected here is well understood by the staff, although its usefulness for tracking human behaviour might vary to a high degree. Not every machine might expose enough information to
  establish models for tracking human behaviour.
  
  \item[\circled{b}] In case not enough data is available, additional special sensors that track  human presence \cite{gollan2014} have to be installed. They do not serve a purpose in the production. These additional sensors contribute additional data, that is very well suited for tracking human behavior, but all the data has to be connected to existed data streams that for the production of individual parts. Serious additional effort for a data-scientist might be required \cite{haslgruebler2019}.  
  
  \item[\circled{c}] If (a) and (b) are not possible,  human object interaction (HOI) detection through cameras \cite{huang2017speed} has to be employed. This is a relatively young and complex field with high  computational efforts, as both human poses and objects have to be detected, and their possible interactions have to deducted, e.g. through reasoning about interaction points \cite{wang2020learning}.
  
  \item[\circled{d}] For establishing a base-line model specialised intrusive sensor equipment \cite{stolte2020,Sztyler2017} might be employed temporary in early phases,
  but not during production.
  
\end{itemize}

In order to fulfil the CASS premise of providing individualized feedback and adaptations, individual humans have to be identifiable. It is thus a prerequisite that the system knows at all times which worker is interacting with machines or the environment. In the scenarios above this
is to a certain extent already the case, as works are tracked through explicit registration procedures for various activities. If this is not the case, establishing additional registrations procedures based on already existing mechanisms should be possible. In this special case even optical tracking of QR codes on mandatory badges is a plausible possibility.

For CASS it is imperative to break down \textbf{human-machine interactions} into fine-grained \textbf{observable interactions} $I$ \cite{haslgruebler2017}. Such interactions $I$ might include: putting raw material into the spindle of a turning machine, mount a tool into a holder, \ldots{}. The more fine-grained the better.

As tracking deviations for each individual $I$ might lead to an over-sensitive system, the system will work based on interaction groups $G$, with $G = \{I_1,\ldots,I_n\}$. Deviations in $G$ will be focused on timing of human-machine interaction and will include:

\begin{itemize}
  \item Delay of groups of tasks.
  \item Erratic timing in groups of tasks.
  \item Timing-drifts in groups of tasks.
\end{itemize}

Independently of these observations a set of potential actions can be established, should specific human behaviour be detected by CASS, it can respond with the following potential actions:
\begin{enumerate}
    \item Change machine properties: the speed of a saw,a conveyor or robot can be changed as long as no functional safety requirements are violated.
    \item Change machine environment: safety distances can be changed on the fly as as long as no functional safety requirements are violated.
    \item Slow down humans: in a production orchestration system implementing for example a worker assistance system the speed at which tasks are communicated to a worker can be varied.
    \item Provide additional information: workers can receive additional information, for example through visual  overlays \cite{funk2016motioneap}, sonification \cite{ballagas2003istuff}, or other means \cite{carter2013ultrahaptics}. For example during the mounting of a tool holder a system might projects dimension information directly onto a work-table, reducing the potential for errors at the cost of a lower work-speed.
\end{enumerate}

For each of the categories above, for each machine a control mechanism has to exist to enact the changes.
Finally the a set of rules has to established that connects actions to observed behavior. As discussed before
the efficiency of the rules can be directly observed by monitoring safety violations and the time required continue
production. For CASS we plan a self-adapting system that learns which rules have been successful, proposes changes, and eventually might be able to decide upon changes thus implementing reinforcement learning.

In order to close the loop, and thus enacting the actions, CASS will employ a Manufacturing Orchestration Systems (MOS), such as \cite{DBLP:conf/bpm/ManglerPRE19}. The MOS will orchestrate all data collection, provide contextualized data-streams (including human behavioral data, quality data, order and work-piece information) to a modular CASS analysis engine. The result will again be collected by the MOS and provided to an modular CASS reasoning engine, that will propose a set of actions. The MOS will then enact the actions and provide feedback to the reasoning functionality. 

\begin{figure}
  \centering
  \includegraphics[width=0.9\textwidth]{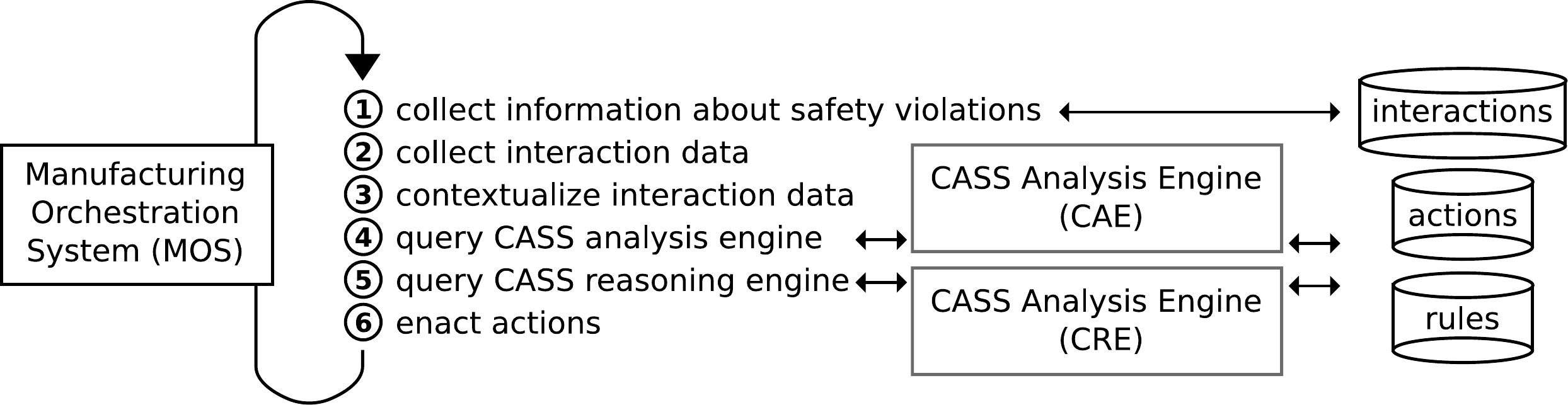}
  \caption{CASS Coordination Architecture}
  \label{fig:architecture}
\end{figure}

This leads to the architecture depicted in Fig. \ref{fig:architecture}.

\section{Evaluation} \label{sec:eval}

Manufacturing companies in high-cost environment are facing continuous pressure to increase productivity, in particular labour productivity to stay competitive in a globalised market~\cite{de2017competitive}. Standard solution during the last decade like outsourcing and offshoring were based on the idea to take advantage of manufacturing in low-cost environment. An unanticipated side effect is an increasing volatility, uncertainty, complexity and ambiguity (VUCA) that may reduce the overall supply chain profitability~\cite{bennett2014vuca}.

Robot systems with cognitive aware safety systems do have the potential to change the “rules of the game”, i.e. to enable different types of manufacturing to be more competitive (e.g., faster, better, more flexible). In a high-cost environment in the future, that will lead to in-sourcing and re-shoring/back-shoring~\cite{lam2016addressing}. The impact measurement of integrated robot systems adoptions and related process modifications is facing challenges in research as well as practice. State of the art investment and costing are facing shortcomings in terms of the real impact because dynamic aspects, indirect effect, environmental and social performance are only slightly considered~\cite{brandenburg2014quantitative}. 

The evaluation will be based on a quantitative performance measurement model  as well as expert interviews for the evaluation of cognitive aware safety systems  

\subsection{Quantitative Performance Measurement Model}

To enable a comprehensive and sustainability evaluation of integrated robot systems with cognitive aware safety system adoptions and related process adoptions, a balancing of economic, environmental and social development
performance is required, i.e. the so-called triple bottom line (profit, planet, and people) ~\cite{elkington1997cannibals}. The objective is to contribute the environment, e.g., with less material waste caused by safety issues, and to improve profitability in parallel, without neglecting the social perspective \cite{elkington25years}. 

An evaluation framework and quantitative model will be developed to analyse the effects of technology innovations and process improvements under consideration of its dynamic behaviour~\cite{jammernegg2007performance}. This approach will improve decision support to selected application fields for cognition aware safety systems under consideration of the triple bottom line. At the strategic and tactical level, rapid modelling , i.e., queuing theory as well as simulation models, is of interest in this context~\cite{reiner2009rapid,reiner2010rapid}. Based on a description of the safety systems, the properties of production requests, the constraints of the production system, and the optimization criteria, problem solving systems will generate optimized configurations of production systems ~\cite{rabta2013hybrid}. Relevant input parameters are basic data, e.g., bill of material, operation times (mean, coefficient of variability), inspection times (mean, coefficient of variability), routings, rework, scrap, setup time, mean time to failure, mean time to repair, worker absenteeism as well as flow data, e.g., demand (mean and coefficient of variation). Performance measures that will be provided are work in process (WIP), flow times (waiting time + operations time), resource (labor and equipment) utilization~\cite{reiner2009rapid,reiner2010rapid}. 

The Cognition Aware Safety Systems might be investigated from a supply chain perspective as well. In this regard the ripple effect might be of special interest. The ripple effect occurs if process disruptions cannot be localized and cascades downstream in the supply chain, e.g. caused by safety issue~\cite{dolgui2018ripple}. Furthermore, the ripple effect can be a driver of the bullwhip-effect~\cite{dolgui2020does}.The bullwhip effect describes the increase of demand volatility upstream in supply chains caused by the lack of coordination~\cite{lee2004comments}. Therefore, performances measure related to the bullwhip effect are of interest because this effect is a cost driver as well, and will increase, e.g., inventory costs, transportation cost, manufacturing cost, replenishment lead time cost, shipping and receiving cost~\cite{chen2012bullwhip}. Overall an increase of the bullwhip effect decrease profitability~\cite{sunil2018supply}.  

This explains why traditional cost-accounting based performance measures facing shortcomings because their focus is too narrow. Furthermore, financial performance measure like return on investment (ROI), EBIT, IRR, Cash Flow, etc. are still very important but can be only a starting point~\cite{kaplan2009conceptual}. 

To be able to better understand the impact of integrated robot systems with cognitive aware safety system it is necessary to develop the dynamic dependencies between the different performance dimensions as well. Therefore, in addition to rapid modeling system dynamics modeling is relevant to treat the complexity of the related processes, i.e., non-linear relationships, delays and closed loops~\cite{sterman2000business,forrester1997industrial}.

\subsection{Expert Interviews}

The following methods, approaches and standards provides the foundation for the expert interview protocol development:
\begin{itemize}
    \item SMED method: setup: factors that affect the decision-making process for selecting the "right" setup technique, i.e, time, cost, energy, facility layout, safety, life, quality and maintenance~\cite{almomani2013proposed}.
    \item TPM: TPM is an approaches to increase stability in the process flow because Well-maintained equipment is facing less breakdowns and generate less quality issues~\cite{wickramasinghe2016effect}, i.e.,. it is mitigating process variability, leading to cost reduction and higher service levels, and efficiency increase~\cite{hooi2017total}. This is represented by a common performance indicator the so called Overall Equipment Effectiveness (OEE), given by the product of availability, performance, and quality indices~\cite{tortorella2021integration}.
    \item ISO 22400: This standard defines a set of key performance indicators (KPI) to evaluate manufacturing processes. These KPIs are described by equations, corresponding parameters, measuring units, etc. The ISO 22400 aims at defining the most important and generally used measures for  manufacturing processes~\cite{zhu2018key}.
    \item SCOR:The Supply Chain Operations Reference (SCOR) model includes five dimensions depicting the SC, i.e, ‘Plan’, ‘Source’, ‘Make’, ‘Deliver’, ‘Return’~\cite{doi:10.1080/09537287.2020.1857874}.The SCOR model is a tool to build standardized, comparable \& measurable processes and related performance measures for supply chain processes~\cite{reiner2006efficiency}.
\end{itemize}
Based on these methods and the related performance measures a survey protocol (guide) has been developed, see~\ref{sec:appendix}. Based on the carefully selected application scenarios we conducted semi-structured interviews with production experts, i.e., we applied an exploratory research design to understand the impact of cognition aware safety systems. The main goal is to get insights base on the knowledge and experience of the experts~\cite{misoch2019qualitative}. Bogner et. al. ~\cite{bogner2009introduction} provide an introduction to the related methodological discussion. Interviews are encouraged to disclose their views and to express related issues. The interviews lasted 60 minutes and in addition secondary data are collected and analysed. Each interview is transcribed and content analysis is applied to the provided textual data. The collected secondary data will be used to enable the analysis based on the developed quantitative performance measurement model.

\section{Conclusion}
\label{sec:conclusion}

So far, all safety related systems in the industry are statically conﬁgured and certiﬁed. Conﬁguration changes after commissioning and safety certification of a machine or even during operation of a production facility are not intended and even prohibited.
To realize the full potential of machinery in smart manufacturing, it is inevitable to adapt functional safety systems in order to enable dynamic reconfiguration. Therefore, a new ﬂexible way of safety conﬁguration, based on the principle of a modular veriﬁcation procedure is needed. Such a Flexible Safety System (FSS) would allow safety adjustments also during operation \cite{etz_flexible_2020}.
This flexible approach opens up the possibility that machinery is able to adapt the functional safety configuration to the skill level of the operator. The combination of FSS together with Cognition Aware Safety Systems (CASS) would enable a machine to detect and calculate the current risk of a hazard, compile an appropriate and pre-certified safety configuration, and deploy the generated configuration.
An example of this scenario is the handling of safety sensors of a maintenance gate on a machine depending on the skill level of the operator. Only maintenance staff should be allowed to open a maintenance gate without activation of safety measures. For all other personnel the machine will be brought into a safe state if the maintenance gate is opened. In order to handle such a feature safely, the machine requires knowledge about skill level and cognitive state of the operator which CASS can provide.

The purpose of this paper is to outline the cognitive as well as the safety foundations that lead to conception of CASS. This paper furthermore details scenarios that will be the basis for a CASS realization, a preliminary CASS architecture, as well as a detailed evaluation strategy to assess the feasibility of the approach.

As detailed in the paper, the realization will proceed in cooperation with our industrial partners. The interdisciplinary nature of the undertaking requires expertise in functional safety, monitoring and analysing cognitive load, data analysis, machine learning methods, and manufacturing orchestration. Thus we will continuously evaluate the progress in order ensure the manifestation of the benefits described in this paper.

\section*{Acknowledgment} This work has been partly funded by the
Austrian Research Promotion Agency (FFG) via the ``Austrian Competence Center
for Digital Production'' (CDP) under the contract number 854187 and the FFG-COMET-K1 Center "Pro$^2$Future" (Products and Production Systems of the Future), Contract No. 881844. This work has been supported by the Pilot Factory Industry 4.0, Seestadtstrasse 27, Vienna,
Austria.

\section{Appendix - Questionnaire}
\label{sec:appendix}

The following questions are to be discussed with technical and managerial experts from the industry, the questions address (1) direct micro economic effect usually expressed in terms of manufacturing related Key Performance Indicators (2) effect related to the effects on employees and (3) aspects related to  mid to long term technology management. 

\subsection{Direct (internal) Economics Effects - Production - KPIs}

\begin{enumerate}
    \item Inhowfar would you expect an improvement of the following production related KPIs according to ISO WD22400/2 due to the implementation of the described solution? Please indicate the respective improvement in percentage of the current values
        \subitem Worker efficiency 
        \subitem Throughput rate (of the [sub]system described in the use case)
        \subitem employee Availability
        \subitem equipment Availability
        \subitem Quality ratio
        \subitem Production process ratio
        \subitem First pass yield
        \subitem Scrap ratio
        \subitem Rework ratio
        \subitem Process capability index
        \subitem Inventory turns
   \item What would be your expectation regarding the effect on the overall equipment effectiveness (OEE)?   
   \item To what extent (\%) do you expect that the described solution can reduce the Cash-to-Cash Cycle Time
   \item What would be the maximum ROI [years] that you would accept for the investment into the described solution?
   \item To what extent can quality issues be eliminated (Reduction of NOK-Parts) from currently x\% to y?\%
   \item According to your work experience can the proposed solution decrease the ratio of scrap parts ?
        \subitem If yes, to what extent (from x\% to y\%) ?
   \item To what extent (\%) do you expect that the described solution can shorten the cycle time?
   \item To what extent (\%) do you expect that the described solution can improve the lead time efficiency
   \item To what extent (\%) do you expect that the described solution can reduce the WIP (work in process)
   \item To what extend (\%) to you think that the described solution will reduce manual working hours for part loading (full repetitive activity)
   \item To what extend (\%) to you think that the described solution will reduce manual working hours for set-up activities (repetitive at the start of a new lot)
   \item How many operating hours per machining centre are currently due to unplanned interruptions (MTBF)?
        \subitem Could the proposed CASS decrease it ?
        \subitem if yes, to what extent ?
   \item To what extent (\%) do you expect that the described solution can reduce the repair time (MTTR)   
   \item What would be the minimum effect in terms of quality improvement (Quality Ratio) and gained operating hours (Throughput rate) so that you would decide in favour of such a system?
   \item What would be your expectation regarding the effect on the overall manufacturing cost?
\end{enumerate}
 
\subsection{Effects related to Employees}

\begin{enumerate}

    \item Do you think that the described solution leads to additional safety risks?, if yes, which ones...
   \item Do you think, the employees would react positively on the implementation of the described use case?
   \item What reactions from the employees would you expect, for what reasons?
   \item What measures would you recommend in order to improve the employee acceptance?
   \item Would you recommend that the approaches of the solution should be considered in regulations pertaining to occupational safety and health administration (OSHA)?
   \item If yes, which aspect would be the most important one:
        \subitem Safety at part loading / unloading
        \subitem Safety during system reconfiguration and/or maintenance?
        \subitem Other, please specify… 
   \item Do you think that the solution would support the transition to the next generations of the workforce (Digital Natives)? 
   \item Do you see difficulties due to the use of AI as a component defining the safety configuration for a particular situation? 
   \item Can you imagine one or more of the following negative effects due to the implementation of the described solution?
        \subitem elimination of cognitive stimuli for the operators - decay of expertise.
        \subitem excess in complexity or dependency on solutions providers.
        \subitem attention "drift” i.e. reliance on the support system increases negligence resulting in a similar number of defects as before
        \subitem "worse" safety situation than before
        \subitem other, please specify
\end{enumerate}
   
   \subsection{Aspects pertaining to Technology Management}
   
 \begin{enumerate}
   \item Have you already considered to invest in (elements of) the described solution? 
   \item Do you know of competitors working into the direction of the described solution? / How many?
   \item On what time horizon can you imagine that the described solution will be adopted by the 10\% most innovative companies in your field? 
   \item Where do you see the most important field of innovation in your field of activity:
        \subitem Data Analytics
        \subitem Flexibility
        \subitem Operator Support (Manual Operations like loading, Unloading, Quality Control)
        \subitem Operator Support (Programming, Planning Operations)
        \subitem Tool Management
        \subitem Management of Fixtures \& Mounting Tools
        \subitem In-Process Measurements
        \subitem Predictive Functions for Quality Assurance
        \subitem Predictive Functions for Maintenance
        \subitem other, please specify
\end{enumerate}

\bibliographystyle{alpha}
\bibliography{intro_bib.bib,tum.bib,safety.bib,wu.bib,various.bib}
\end{document}